\documentclass[a4paper]{article}

\usepackage{INTERSPEECH2020}
\usepackage{mathrsfs}
\usepackage{subfigure}
\usepackage{graphics}
\usepackage{multirow}
\usepackage{multicol}
\usepackage{amsmath}
\usepackage{booktabs}       
\usepackage{amsfonts}       
\usepackage{nicefrac}       
\usepackage{microtype}
\usepackage{float}
\usepackage[misc]{ifsym} 

\title{S2IGAN: Speech-to-Image Generation via Adversarial Learning}
\name{Xinsheng Wang$^{1,2}$, Tingting Qiao$^{2,3}$, Jihua Zhu$^1$ $^{(\textrm{\Letter})}$, Alan Hanjalic$^2$, Odette Scharenborg$^2$}
\address{
  $^1$School of Software Engineering, Xi'an Jiaotong University, China.\\
  $^2$Multimedia Computing Group, Delft University of Technology, Delft, The Netherlands.\\
$^3$College of Computer Science and Technology, Zhejiang University, China.}
\email{wangxinsheng@stu.xjtu.edu.cn, qiaott@zju.edu.cn, zhujh@xjtu.edu.cn, a.hanjalic@tudelft.nl, o.e.scharenborg@tudelft.nl}

\begin{document}

\maketitle
\begin{abstract}
An estimated half of the world's languages do not have a written form, making it impossible for these languages to benefit from any existing text-based technologies. In this paper, a speech-to-image generation (S2IG) framework is proposed which translates speech descriptions to photo-realistic images without using any text information, thus allowing unwritten languages to potentially benefit from this technology. The proposed S2IG framework, named S2IGAN, consists of a speech embedding network (SEN) and a relation-supervised densely-stacked generative model (RDG). SEN learns the speech embedding with the supervision of the corresponding visual information. Conditioned on the speech embedding produced by SEN, the proposed RDG synthesizes images that are semantically consistent with the corresponding speech descriptions. Extensive experiments on datasets CUB and Oxford-102 demonstrate the effectiveness of the proposed S2IGAN on synthesizing high-quality and semantically-consistent images from the speech signal, yielding a good performance and a solid baseline for the S2IG task.

\end{abstract}


\section{Introduction}
The recent development of deep learning and Generative Adversarial Networks (GAN) \cite{goodfellow2014generative,mirza2014conditional,balaji2019conditional} 
led to many efforts being carried out on the task of image generation conditioned on natural languages \cite{reed2016generative,zhang2018stackgan++,xu2018attngan,qiao2019mirrorgan,yin2019semantics,tan2019semantics}. Although great progress has been made, most of the existing natural language-to-image generation systems use text descriptions as their input, also referred to as Text-to-Image Generation (T2IG). Recently, a speech-based task was proposed in which face images are synthesized conditioned on speech \cite{oh2019speech2face,wen2019face}. This task, however, only considers the acoustic properties of the speech signal, but not the language content. Here, we present a natural language-to-image generation system that is based on a spoken description, bypassing the need for text. We refer to this new task as Speech-to-Image Generation (S2IG). This is similar to the recently proposed task of speech-to-image translation task \cite{li2020direct}.

This work is motivated by the fact that an estimated half of the 7,000 languages in the world do not have written forms \cite{lewis2015ethnologue} (so-called unwritten languages), which makes it impossible for these languages to benefit from any existing text-based technologies, including text-to-image generation. The Linguistic Rights as included in the Universal Declaration of Human Rights state that it is a human right to communicate in one’s native language. For these unwritten languages, it is essential to develop a system that bypasses text and maps speech descriptions to images. Moreover, even though existing knowledge and methodology make `speech2text2image' transfer possible, directly mapping speech to images might be more efficient and straightforward. 

In order to synthesize plausible images based on speech descriptions, speech embeddings that carry the details of semantic information in the image need to be learned. To that end, we decompose the task of S2IG into two stages, i.e., a speech semantic embedding stage and an image generation stage. Specifically, the proposed speech-to-image generation model via adversarial learning (which we refer to as S2IGAN) consists of a Speech Embedding Network (SEN), which is trained to obtain speech embeddings by modeling and co-embedding speech and images together, and a novel Relation-supervised Densely-stacked Generative Model (RDG), which takes random noise and the speech embedding embedded by SEN as input to synthesize photo-realistic images in a multi-step (coarse-to-fine) way. 

In this paper, we present our attempt to generate images directly from the speech signal bypassing text. This task requires specific training material consisting of speech and image pairs. Unfortunately, no such database, with the right amount of data, exists for an unwritten language. The results for our proof-of-concept are consequently presented on two databases with English descriptions, i.e., CUB \cite{wah2011caltech} and Oxford-102 \cite{nilsback2008automated}. The benefit of using English as our working language is that we can compare our S2IG results to T2IG results in the literature. Our results are also compared to those of \cite{li2020direct}.

\section{Approach}
Given a speech description, our goal is to generate an image that is semantically aligned with the input speech. To this end, S2IGAN consists of two modules, i.e., SEN to create the speech embeddings and RDG to synthesize the images using these speech embeddings.

\begin{figure*}[ht]
    \centering
    \includegraphics[width=0.9\textwidth]{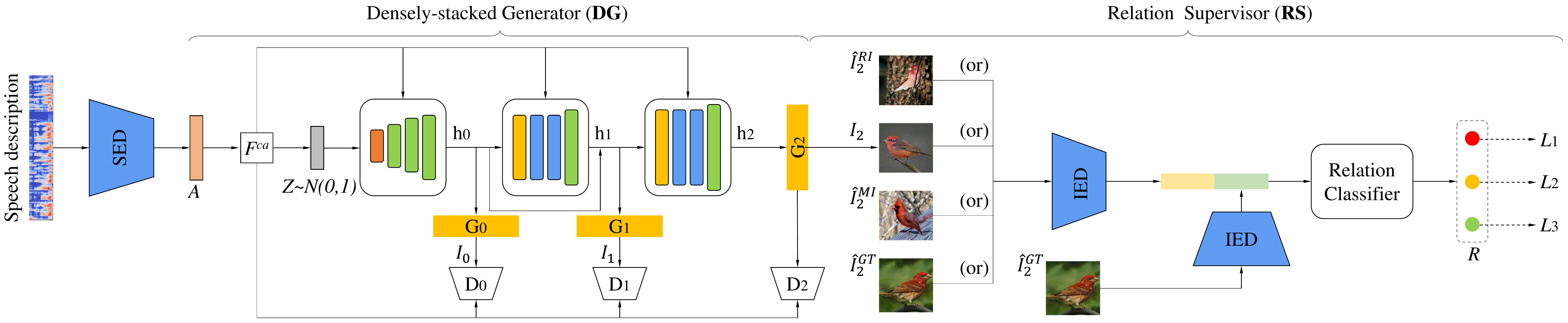}
    \caption{Framework of the relation-supervised densely-stacked generative model (RDG). ${\hat I}^{RI}_2$ represents a real image from the same class as the ground-truth image (${\hat I}^{GT}_2$), ${I}_2$ represents a fake image synthesized by the framework. ${\hat I}^{MI}_2$ represents a real image from a different class as ${\hat I}^{GT}_2$. $L_i$ indicates labels for three types of relations. SED and IED are pre-trained in SEN.}
    \label{fig:network_gan}
\end{figure*}

\subsection{Datasets}
CUB \cite{wah2011caltech} and Oxford-102 \cite{nilsback2008automated} are two commonly-used datasets in the field of T2IG \cite{reed2016generative,zhang2018stackgan++}, and were also adopted in the most recent S2IG work \cite{li2020direct}. CUB is a fine-grained bird dataset that contains 11,788 bird images belonging to 200 categories and Oxford-102 is a fine-grained flower dataset contains 8,189 images of flowers from 102 different categories. Each image in both datasets has 10 text descriptions collected by \cite{reed2016learning}. Since there are no speech descriptions available for both datasets, we generated speech from the text descriptions using tacotron2 \cite{shen2018natural} which is a text-to-speech system\footnote{https://github.com/NVIDIA/tacotron2}.

\subsection{Speech Embedding Network (SEN)}

Given an image-speech pair, SEN tries to find a common space for both modalities, so that we can minimize the modality gap and obtain visually-grounded speech embeddings. SEN is a dual encoder framework, including an image encoder and a speech encoder, which is similar to the model structure in \cite{merkx2019language}. 

\textbf{\textit{The image encoder (IED)}} adopts the Inception-v3 \cite{szegedy2016rethinking} pre-trained on ImageNet \cite{russakovsky2015imagenet} to extract visual features. On top of it, a single linear layer is employed to convert the visual feature to a common space of visual and speech embeddings. As a result, we obtain an image embedding $V$ from IED.

\textbf{\textit{The speech encoder (SED)}} employs a structure similar to that of \cite{merkx2019language}. Specifically, it consists of a two-layer 1-D convolution block, two-layer bi-directional gated recurrent units (GRU) \cite{cho2014learning} and a self-attention layer. Finally, speech is represented by a speech embedding $A$ in the common space. The input of the SED are log Mel filter bank spectrograms, which are obtained from the speech signal using 40 Mel-spaced filter banks with 25 ms Hamming window and 10 ms shift.

More details of SEN, including the framework illustration, can be found on the project website\footnote{For more details on the model and results, please see: https://xinshengwang.github.io/project/s2igan/\label{ft:project}}.

\subsubsection{Objective Function}
To minimize the distance between a matched pair of an image feature and speech feature while maintaining discrimination of the features compared to features from other bird (CUB) or flower (Oxford-102) classes, matching loss and distinctive loss are proposed.

\textbf{\textit{Matching loss}} is designed to minimize the distance of a matched image-speech pair. Specifically, in a batch of image-speech embedding pairs $\left\{ {\left( {{V_i},{A_i}} \right)} \right\}_i^n$, where $n$ is the batch size, the probability for the speech embedding $A_i$ matching with the image embedding $V_i$ is
\begin{equation}
\label{eq:A-V_matching}
\small
   P\left( {{V_i}|{A_i}} \right) = \frac{{\exp \left( {\beta S\left( {{A_i},{V_i}} \right)} \right)}}{{\sum\nolimits_{j = 1}^n {{M_{i,j}}\exp \left( {\beta S\left( {{A_i},{V_j}} \right)} \right)} }},
\end{equation}
where $\beta$ is a smoothing factor, set as 10 following \cite{xu2018attngan}. $S\left( {{A_i},{V_i}} \right)$ is a cosine similarity score of ${A_i}$ and ${V_i}$. As in a mini-batch, we only treat $\left( {{V_i},{A_i}} \right)$ as a positive matched pair, therefore we use a mask $M_{i,j} \in \mathbb{R}^{n\times n}$ to deactivate the effect of pairs from the same class. Specifically,
\begin{equation}
\small 
{M_{i j}}=\left\{\begin{array}{ll}
{0,} & {\text { if } A_{i} \text { matches } V_{j}}\; { \& } \; {i \ne j}, \\
{1,} & {\text { otherwise }},
\end{array}\right.
\end{equation}
where $A_i$ matches $V_j$ means they come from the same class. The loss function is then defined as the negative log probability of $P\left( {{V_i}|{A_i}} \right)$:
\begin{equation}
\small
    {{\cal L}_{A - V}} =  - \sum\limits_{i = 1}^n {\log } P\left( {{V_i}|{A_i}} \right).
\end{equation}
Reversely, we also calculate ${{\cal L}_{V - A}}$ for $V_i$ matching $A_i$. The matching loss is then calculated as
\begin{equation}
   {\cal L}_m = {{\cal L}_{A - V}} + {{\cal L}_{V - A}}.
\end{equation}

\textbf{\textit{Distinctive loss}} is designed to ensure that the space is optimally discriminative regarding the instance classes. Specifically, both speech and image features in the embedding space are converted to a label space by adding a perception layer, i.e., ${\hat V_i} = f\left( {{V_i}} \right)$ and ${\hat A_i} = f\left( {{A_i}} \right)$, where $\hat V_i, \hat A_i \in \mathbb{R}^{N}$ and $N$ is the number of classes. The loss function is given by
\begin{equation}
\small
    {\cal L}_d =  - \sum\limits_{i = 1}^n {\left( {\log \hat P\left( {{C_i}|{{\hat A}_i}} \right) + \log \hat P\left( {{C_i}|{{\hat V}_i}} \right)} \right)}, 
\end{equation}
where ${\hat P( {{C_i}|{{\hat A}_i}} )}$ and ${\hat P( {{C_i}|{{\hat V}_i}} )}$ represent softmax probabilities for ${{\hat A}_i}$ and ${{\hat V}_i}$ belonging to their corresponding class ${C_i}$.

\textbf{\textit{Total loss}} for training SEN is finally given by
\begin{equation}
\label{eq:distinctive_loss}
\small
    {{\cal L}_{SEN}} = {{\cal L}_m} + {{\cal L}_d}.
\end{equation}

\subsection{Relation-supervised Densely-stacked Generative Model (RDG)}
After learning the visually-grounded and class-discriminative speech embeddings, we employ RDG to generate images conditioned on these speech embeddings. RDG consists of two sub-modules, which are a Densely-stacked Generator (DG) and a Relation Supervisor (RS), see Figure \ref{fig:network_gan}. 
\subsubsection{Densely-stacked Generator (DG)}
RDG uses the multi-step generation structure \cite{zhang2018stackgan++,qiao2019mirrorgan,yin2019semantics} because of its previously shown performance. This structure generates images from small scale (low-resolution) to large scale (high-resolution) step by step. Specifically, in our model, ${\rm{64}} \times {\rm{64}}$, ${\rm{128}} \times {\rm{128}}$, and ${\rm{256}} \times {\rm{256}}$ pixel images were generated in multi-steps. To fully exploit the information of the hidden feature $(h_i)$ of each step, we design a densely-stacked generator. With the speech embedding $A$ as input, the generated image in each stacked generator can be expressed as follows: 
\begin{equation}
\small
    \begin{array}{*{20}{l}}
{{h_0} = {F_0}\left( {z,{F^{ca}}(A)} \right),}\\
{{h_i} = {F_i}\left( {{h_0}, \ldots ,{h_{i - 1}},{F^{ca}}(A)} \right),i \in \left\{ {1,2} \right\},}\\
{{I_i} = {G_i}\left( {{h_i}} \right),i \in \left\{ {0,1,2} \right\}}, 
\end{array}
\end{equation}
where $z$ is a noise vector sampled from a  normal distribution. $F^{ca}$ represents Conditioning Augmentation \cite{zhang2017stackgan,zhang2018stackgan++} that augments the speech features thus produces more image-speech pairs. It is a popular and useful strategy which is used in most recent text-to-speech generation tasks \cite{tan2019semantics,xu2018attngan,qiao2019mirrorgan}. $h_i$ is the hidden feature from the non-linear transformation $F_i$. $h_i$ is fed to the generator $G_i$ to obtain image $I_i$. 

\subsubsection{Relation Supervisor (RS)}
To ensure that the generator produces high-quality images that are semantically aligned with the spoken description, we propose a relation supervisor to provide a strong relation constraint to the generation process. Specifically, we form an image set for each generated image $I_i$, i.e., $\{ I_i,  {\hat I}_i^{GT}, {\hat I}_i^{RI}, {\hat I}_i^{MI}\}$ indicating the generated fake image, the ground-truth image, a real image from the same class as $I_i$, and a real image from a different randomly-sampled class, respectively. We then define three types of relation classes: 1) a positive relation $L_1$, between ${\hat I}_i^{GT}$ and ${\hat I}_i^{RI}$; 2) a negative relation $L_2$, between ${\hat I}_i^{GT}$ and ${\hat I}_i^{MI}$; 3) an undesired relation $L_3$, between ${\hat I}_i^{GT}$ and ${\hat I}_i^{GT}$. A relation classifier is trained to classify these three relations. We expect the relation between $I_i$ and ${\hat I}_i^{GT}$ to be close to the positive relation $L_1$, because $I_i$ should semantically align with its corresponding ${\hat I}_i^{GT}$, however, it should not be identical to ${\hat I}_i^{GT}$ to ensure the diversity of the generated results. Therefore, the loss function for training the RS is defined as:
\begin{equation}
\small
\begin{array}{l}
\begin{aligned}
{L_{RS}} = & - \sum\limits_{j = 1}^3 {\log \hat P\left( {{L_j}\left| {{R_j}} \right.} \right)} - \log \hat P\left( {{L_1}\left| {{R_{GT-FI}}} \right.} \right),
\end{aligned}
\end{array}
\end{equation}
where $R_j$ is a relation vector produced by RS with the input of a pair of images with relation $L_j$, e.g., ${R_1} = RS \big( {\hat I}^{GT},{{\hat I}^{RI}} \big)$. ${{R_{GT-FI}}}$ is the vector of relation between ${\hat I}_i^{GT}$ and $I_i$. Note that we apply RS only to the last generated image, i.e., $i=2$, for computational efficiency.

\subsubsection{Objective Function}
The final objective function of RDG is defined as:
\begin{equation}
\small
{{\cal L}_{{G}}}  = \sum_{i=0}^2 {\cal L}_{G_i} + {\cal L}_{RS},
\end{equation}
where the loss function for the $i$-$th$ generator $G_i$ is defined as:
\begin{equation}
\small
\begin{array}{l}
\begin{aligned}
{{\cal L}_{{G_i}}} = & - {\mathbb{E}_{{I_i} \sim {p_G}_i}}\left[ {\log {D_i}\left( {{I_i}} \right)} \right] + \\
 &- {\mathbb{E}_{{I_i} \sim {p_G}_i}}\left[ {\log \left( {{D_i}\left( {{I_i},{F^{ca}\big(A\big)}} \right)} \right)} \right].
 \end{aligned}
\end{array}
\end{equation}
The loss function for the corresponding discriminator $D$ of RDG is given by:
\begin{equation}
\small
{{\cal L}_{{D}}}  = \sum_{i=0}^2 {\cal L}_{D_i},
\end{equation}
where the loss function for the $i$-$th$ discriminator $D_i$ is given by:
\begin{equation}
\small
\begin{array}{l}
\begin{aligned}
{{\cal L}_{{D_i}}} =& - {\mathbb{E}_{{{\hat I}_i} \sim {p_{dat{a_i}}}}}\big[ {\log {D_i}\big( {{{\hat I}_i}} \big)} \big] + \\
&- {\mathbb{E}_{{I_i} \sim {p_G}_i}}\big[ {\log \big( {1 - {D_i}\big( {{I_i}} \big)} \big)} \big] + \\
& - {\mathbb{E}_{{{\hat I}_i} \sim {p_{dat{a_i}}}}}\big[ {\log {D_i}\big( {{{\hat I}_i},{F^{ca}\big(A\big)}} \big)} \big] + \\
& - {\mathbb{E}_{{I_i} \sim {p_G}_i}}\big[ {\log \big( {1 - {D_i}\big( {{I_i},{F^{ca}\big(A\big)}} \big)} \big)} \big].
\end{aligned}
\end{array}
\end{equation}
Here, the first two items are unconditional loss that discriminate the fake and real images, and the last two items are conditional loss discriminating whether the image and the speech description match or not. The $I_i$ is from the model distribution $G_i$ at the $i^{th}$ scale, and ${\hat I}_i$ is from the real image distribution $p_{data_i}$ at the same scale. The generators and discriminators were trained alternately.

\subsection{Evaluation Metrics}
We use two metrics to evaluate the performance of our SI2GAN model. To evaluate \textit{\textbf{diversity and quality}} of the generated images, we used two popular evaluation metrics for quantitative evaluation of generative models as that in \cite{zhang2018stackgan++}:  Inception score (IS) \cite{salimans2016improved} and fr\'{e}chet inception distance (FID) \cite{heusel2017gans}, where, a higher IS means more diversity and a lower FID means a smaller distance between the generated and real image distributions, which indicates better generated images.

The \textit{\textbf{visual-semantic consistency}} between the generated images and their speech descriptions is evaluated through a content-based image retrieval experiment between the real images and the generated images, and evaluated using mAP scores. Specifically, we randomly chose two real images from each class of the test set, resulting in a query pool. Then we used these query images to retrieve generated fake images that belong to their corresponding classes. We used the pre-trained Inception-v3 to extract features of all images. Higher mAP indicates a closer feature distance between fake images and their ground truth images, which indirectly shows a higher semantic consistency between generated images and their corresponding speech descriptions. 

\section{Results}

\begin{table}[tb]
\scriptsize
\centering
\caption{Performance of S2IGAN compared to other methods. $\dagger$ means that the results are taken from the original paper. The best performance is shown in bold.}
\setlength{\tabcolsep}{0.7mm}
\label{tab:mainresult}
\begin{tabular}{lccccccc}
\toprule
& &\multicolumn{3}{c}{CUB (Bird)} & \multicolumn{3}{c}{Oxford-102 (Flower)} \\ \cmidrule(r){3-5} \cmidrule(r){6-8} 
Evaluation Metric &Input & mAP & FID & IS & mAP & FID & IS \\ \midrule
StackGAN-v2 & text & 7.01 & 20.94 & 4.02$\pm$0.03 & 9.88 & 50.38 & 3.35$\pm$0.07 \\
MirrorGAN$\dagger$ & text & --- & --- & 4.56$\pm$0.05 & --- & --- & \--- \\
SEGAN$\dagger$ & text & --- & --- & \textbf{4.67$\pm$0.04} & --- & --- & \--- \\ 
\midrule
\cite{li2020direct}$\dagger$ & speech & --- & 18.37 & 4.09$\pm$0.04 & ---  & 54.76 & 3.23$\pm$0.05 \\ 
StackGAN-v2 & speech & 8.09 & 18.94 & 4.14$\pm$0.04 & 12.18 & 54.33 & \textbf{3.69$\pm$0.08} \\
S2IGAN & speech & \textbf{9.04} & \textbf{14.50} & 4.29$\pm$0.04 & \textbf{13.40} & \textbf{48.64} & 3.55$\pm$0.04 \\ \bottomrule
\end{tabular}
\end{table}

\subsection{Objective Results} 
We compare our results with several state-of-the-art T2IG methods, including StackGAN-v2 \cite{zhang2018stackgan++}, MirrorGAN \cite{qiao2019mirrorgan} and SEGAN \cite{tan2019semantics}. StackGAN-v2 is a strong baseline for the T2IG task and provides the effective stacked structure for the following methods. Both MirrorGAN and SEGAN are based on the stacked structure. MirrorGAN utilizes word-level \cite{xu2018attngan} and sentence-level attention mechanisms, and a ``text-to-image-to-text'' structure for T2IG, and SEGAN also uses word-level attention with extra proposed attention regularization and a siamese structure. In order to allow for a direct comparison on the S2IG task to StackGAN-v2, we reimplemented StackGAN-v2 and replaced the text embedding with our speech embedding. Moreover, we compare our results to the recently released speech-based model by \cite{li2020direct}.

The results are shown in Table \ref{tab:mainresult}. First, our method outperformed \cite{li2020direct} on all evaluation metrics and datasets. Compared with the StackGAN-v2 that took our speech embedding as input, our S2IGAN also achieved higher mAP and lower FID on both datasets. These results indicate that our method is effective in generating high-quality and semantically consistent images on the basis of spoken descriptions. The comparison of our S2IGAN with three state-of-the-art T2IG methods show that the S2IGAN method is competitive, and thus establishes a solid new baseline for the S2IG task. 

Speech input is generally considered to be more difficult to deal with than text because of its high variability, its long duration, and the lack of pauses between words. Therefore, S2IG is more challenging than T2IG. However, the comparison of the performances of StackGAN-v2 on the S2IG and T2IG tasks shows that StackGAN-v2 generated better images using speech embeddings learned by our SEN. Moreover, the StackGAN-v2 based on our learned speech embeddings outperforms \cite{li2020direct} on almost all evaluation metrics and datasets, except for the slightly higher FID on CUB dataset. Note that \cite{li2020direct} takes the native StackGAN-v2 as the generator, which means that the only difference between \cite{li2020direct} and the speech-based StackGAN-v2 in Table \ref{tab:mainresult} is the speech embedding method. These results confirm that our learned speech embeddings are competitive compared to text input and the speech embeddings in \cite{li2020direct}, showing the effectiveness of our SEN module.

\subsubsection{Subjective Results} 

\begin{figure}[htb]
    \centering
    \includegraphics[width=0.85\linewidth]{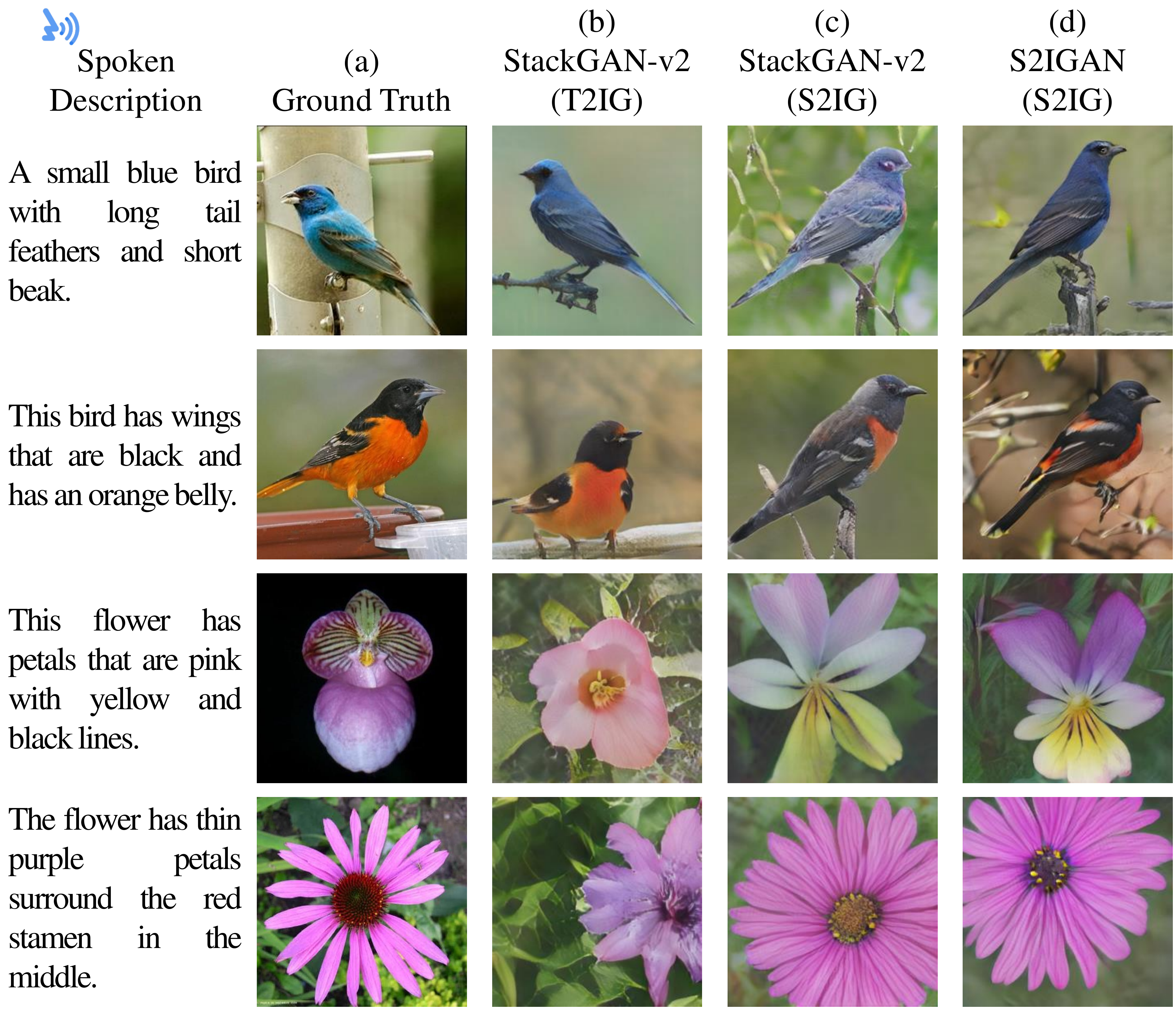}
    \caption{Examples of images generated by different methods.}
    \label{fig:mainresult}
\end{figure}

The subjective visual results are shown in Figure \ref{fig:mainresult}. As can be seen, the images synthesized by our S2IGAN (d) are photo-realistic and convincing. By comparing the images generated by (d) S2IGAN and (c) StackGAN-v2 conditioned on speech embeddings, we can see that the images generated by S2IGAN are clearer and sharper, showing the effectiveness of the proposed S2IGAN on synthesizing visually high-quality images. The comparison of StackGAN-v2 conditioned on (b) text and (c) speech features embedded by the proposed SEN shows that our learned speech embeddings are competitive compared with the text features embedded by StackGAN-v2, showing the effectiveness of SEN. More results are shown on the project website\textsuperscript{\ref{ft:project}}.

\begin{figure}[t]
    \centering
    \includegraphics[width=0.61\linewidth]{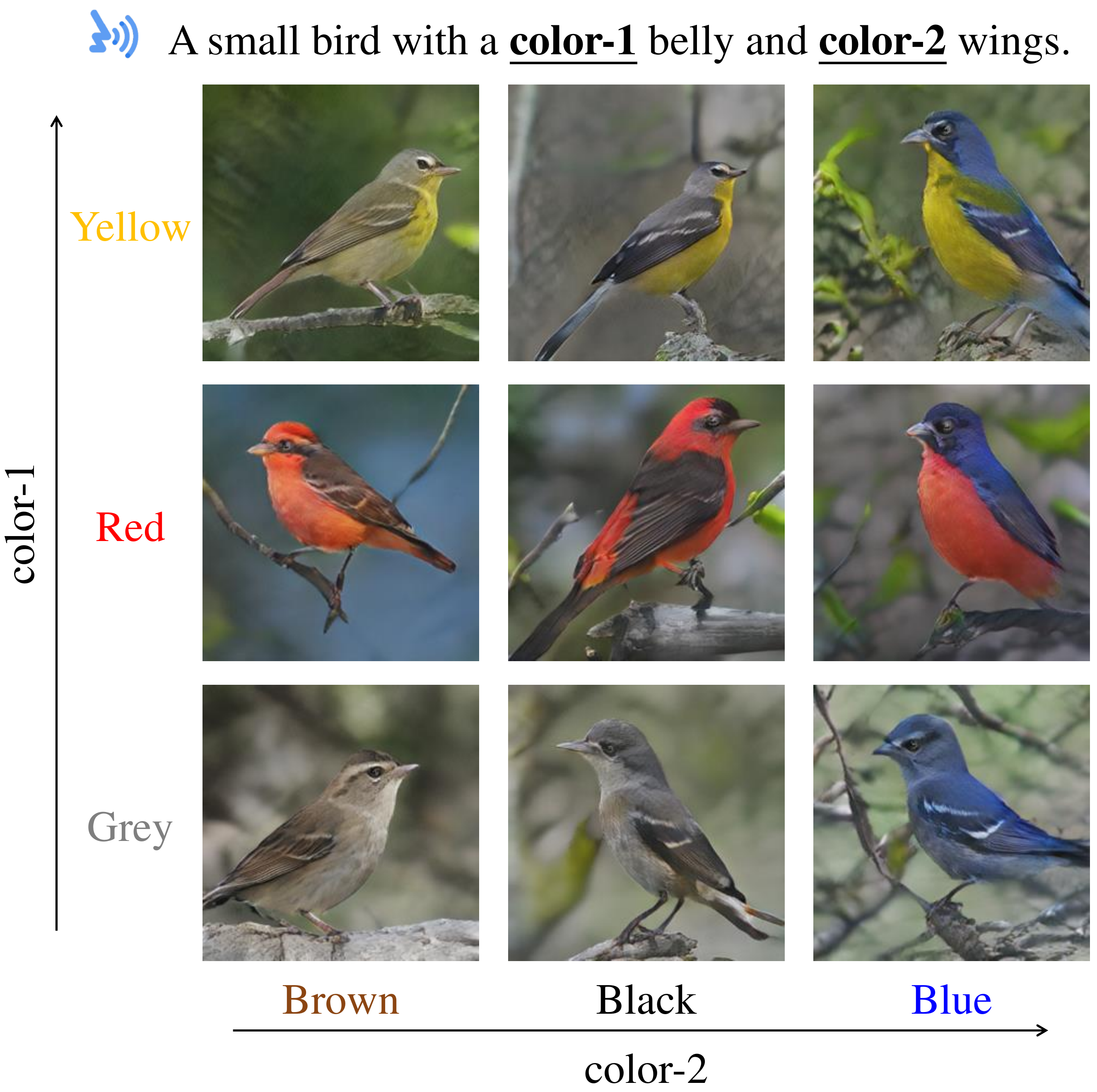}
    \caption{Generated examples by S2IGAN. The generated images are based on speech descriptions with different color keywords.}
    \label{fig:semanticvisual}
\end{figure}

To further illustrate S2IGAN's ability to catch subtle semantic differences in the speech descriptions, we generated images conditioned on speech descriptions in which color keywords were changed. As Figure \ref{fig:semanticvisual} shows, the visual semantics of the generated birds, specifically, the colors of the belly and the wings, are consistent with the corresponding semantic information in the spoken descriptions. These visualization results indicate that SEN successfully learned the semantic information in the speech signal, and that our RDG is capable of capturing these semantics and generating discriminative images that are semantically aligned with the input speech. 

\subsection{Component analysis}

An extensive ablation study investigated the effectiveness of key components of SI2GAN. Specifically, the effects of the densely-stacked structure of DG, RS, and SEN were investigated by removing each of these components respectively. Removing any component resulted in a clear decrease of the generation performance, showing the effectiveness of each component. Details can be found on the project website\textsuperscript{\ref{ft:project}}.

\section{Discussion and Conclusion}
This paper introduces a novel speech-to-image generation (S2IG) task and we developed a novel generative model, called S2IGAN, which tackles S2IG in two steps. First, semantically discriminative speech embeddings are learned by a speech embedding network. Second, high-quality images are generated on the basis of the speech embeddings. The results of extensive experiments show that our S2IGAN has state-of-the-art performance, and that the learned speech embeddings capture the semantic information in the speech signal.

The current work is based on synthesized speech, which makes the current S2IG baseline an upper-bound baseline. The future work will focus on several directions. First, we will investigate this task with natural speech instead of synthesized speech. Second, it will be highly interesting to test the proposed methodology on a true unwritten language rather than the well-resourced English language. Third, we will further improve our methods in terms of efficiency and accuracy, for example, by making end-to-end training more effective and efficient and by applying attention mechanisms to our generator to further improve the quality of the generated images. An interesting avenue for future research would be to automatically discover speech units based on corresponding visual information from the speech signal \cite{harwath2019towards} to segment the speech signal. This would allow us to use segment- and word-level attention mechanisms, which have shown to lead to improved performance on the text-to-image generation task \cite{xu2018attngan}, to improve the performance of speech-to-image generation.  

\section{Acknowledgements}
This work has been partially supported by the China Scholarship Council (CSC).

\bibliographystyle{IEEEtran}
\bibliography{Interspeech2020.bbl}

\end{document}